\documentclass[sigconf]{acmart}
\acmSubmissionID{532}

\usepackage{booktabs} % For formal tables
\usepackage{makecell}
\usepackage{subfigure}
\usepackage{graphicx}
\usepackage[misc]{ifsym}

\def\eg{\textit{e.g.,~}}               % for example
\def\ie{\textit{i.e.,~}}      % that is, in other words
\def\sysname{DreamEditor}

\def\dreambooth3d{D-DreamFusion*}
\def\nerf2nerf{Instruct-N2N}

\def\ourvote{81.1}
\usepackage[ruled]{algorithm2e} % For algorithms

\SetAlFnt{\small}
\SetAlCapFnt{\small}
\SetAlCapNameFnt{\small}
\SetAlCapHSkip{0pt}

\copyrightyear{2023}
\acmYear{2023}
\setcopyright{acmlicensed}\acmConference[SA Conference Papers '23]{SIGGRAPH Asia 2023 Conference Papers}{December 12--15, 2023}{Sydney, NSW, Australia}
\acmBooktitle{SIGGRAPH Asia 2023 Conference Papers (SA Conference Papers '23), December 12--15, 2023, Sydney, NSW, Australia}
\acmPrice{15.00}
\acmDOI{10.1145/3610548.3618190}
\acmISBN{979-8-4007-0315-7/23/12}

\begin{document}
% Title portion
\title{DreamEditor: Text-Driven 3D Scene Editing with Neural Fields}

\author{Jingyu Zhuang*}
\email{zhuangjy6@mail2.sysu.edu.cn}
\affiliation{%
  \institution{Sun Yat-sen University}
  \country{China}
}

\author{Chen Wang*}
\email{cw.chenwang@outlook.com}
\affiliation{%
  \institution{University of Pennsylvania}
  \country{USA}
}

\author{Liang Lin \textsuperscript{\Letter}}
\email{linliang@ieee.org}
\affiliation{%
  \institution{Sun Yat-sen University}
  \country{China}
}

\author{Lingjie Liu \textsuperscript{\Letter}}
\email{lingjie6@seas.upenn.edu}
\affiliation{%
  \institution{University of Pennsylvania}
  \country{USA}
}

\author{Guanbin Li \textsuperscript{\Letter}}
\email{liguanbin@mail.sysu.edu.cn}
\affiliation{%
  \institution{Sun Yat-sen University}
  \country{China}
}

\thanks{
*Both authors contributed equally to this research. Corresponding authors: Guanbin Li, Lingjie Liu and Liang Lin. 
Welcome to \href{https://github.com/zjy526223908/DreamEditor}{\textcolor{red}{\emph{Code}}} and \href{https://www.sysu-hcp.net/projects/cv/111.html}{\textcolor{red}{\emph{Project page}}}
}

\renewcommand{\shortauthors}{Zhuang et al.}

\begin{abstract}

Neural fields have achieved impressive advancements in view synthesis and scene reconstruction. However, editing these neural fields remains challenging due to the implicit encoding of geometry and texture information. In this paper, we propose \sysname{}, a novel framework that enables users to perform controlled editing of neural fields using text prompts. 
By representing scenes as mesh-based neural fields, \sysname{} allows localized editing within specific regions. 
\sysname{} utilizes the text encoder of a pretrained text-to-Image diffusion model to automatically identify the regions to be edited based on the semantics of the text prompts. Subsequently,  \sysname{} optimizes the editing region and aligns its geometry and texture with the text prompts through score distillation sampling~\cite{poole2022dreamfusion}. Extensive experiments have demonstrated that \sysname{} can accurately edit neural fields of real-world scenes according to the given text prompts while ensuring  consistency in irrelevant areas. \sysname{} generates highly realistic textures and geometry, significantly surpassing previous works in both quantitative and qualitative evaluations.
\end{abstract}



\begin{teaserfigure}
  \centering
  \includegraphics[width=\textwidth]{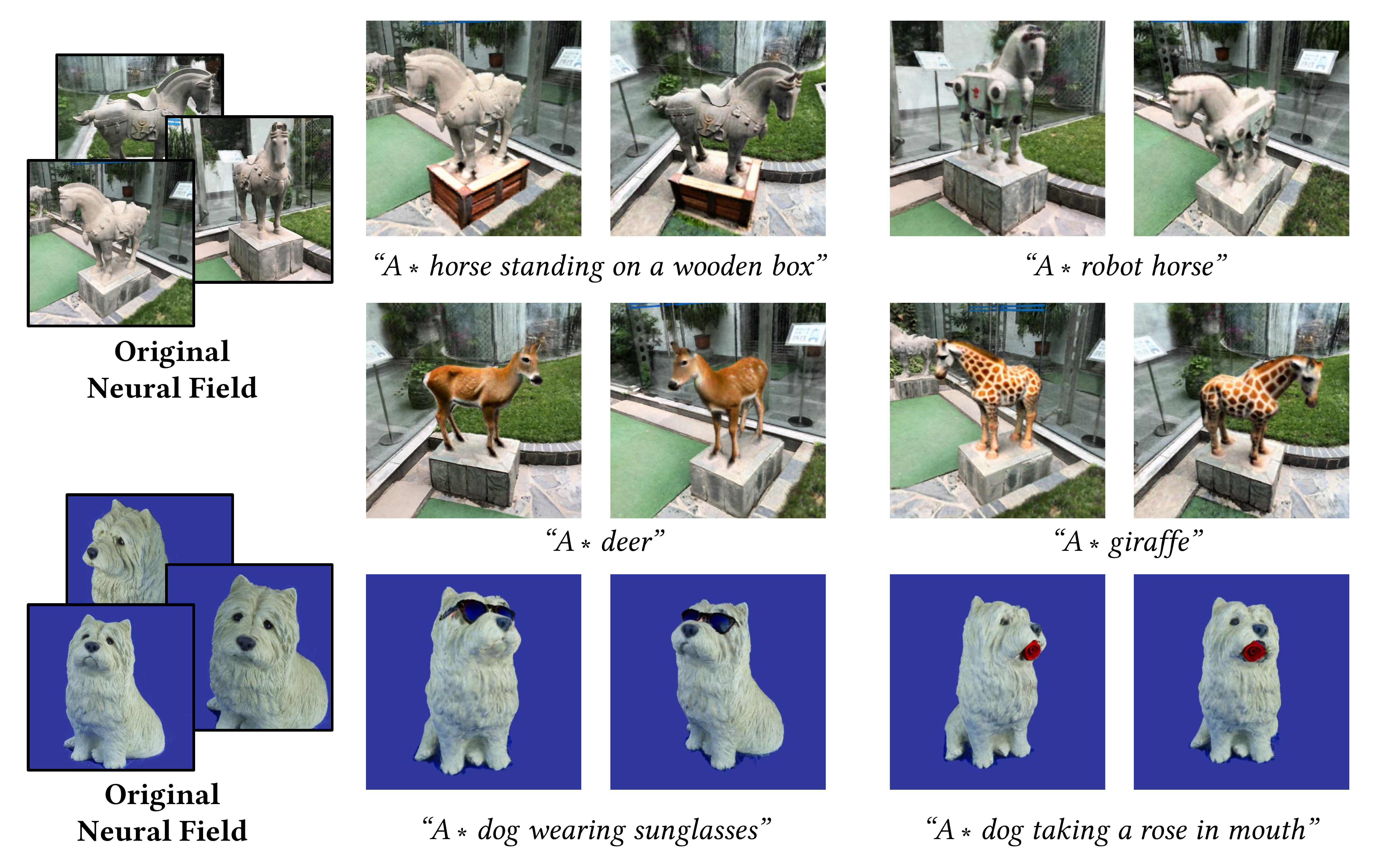}
  \caption{Our approach \sysname{} allows users to edit 3D scenes with text prompts. \sysname{} achieves precise and high-quality editing that maintains irrelevant regions unchanged.}
  \label{fig:teaser}
\end{teaserfigure}

\maketitle

% \begin{figure*}
%   \includegraphics[width=\textwidth]{pic/1.png}
%   \caption{We propose \sysname{}, a NeRF editing framework that allows users to modify NeRF using text prompts. The images in solid blocks represent the rendered images of the edited NeRF, while those in dotted blocks depict the extracted meshes.}
%   \label{fig_begin}
% \end{figure*}

\section{Introduction}
% Para 1: The development of NeRF
Neural radiance fields ~\cite{mildenhall2021nerf}, NeuS~\cite{wang2021neus} and subsequent research~\cite{ liu2020neural, muller2022instant, wang2022nerf} (collectively referred to as \textit{neural fields}) have made significant progress in scene reconstruction and novel view synthesis.
By capturing multi-view images of a 3D scene and using off-the-shelf structure-from-motion models to estimate camera poses, one can train neural networks to learn neural fields that implicitly represent the geometry and texture of the scene. 
% Neural fields are not only capable of synthesizing high-quality renderings at novel viewpoints but can also be reconstructed to 3D triangular meshes with marching cubes~\cite{lorensen1987marching} or specialized weighting functions~\cite{wang2021neus, yariv2021volume}. 
Compared to the traditional pipeline involving tedious 3D matching and complex postprocessing steps, neural fields offer a more efficient and accessible approach for reconstructing real-world objects and scenes into Computer Graphics assets for general users.

% Para 2: mention the problem of editing NeRF 1. without text is not convenient 复杂的操作 2. given text is without locating
However, editing neural fields is not a straightforward task since the shape and texture information is implicitly encoded within high-dimensional neural network features. Conventional 3D modeling techniques are ineffective for manual sculpting and re-texturing since explicit geometry is not available.
Previous research has explored techniques for neural fields editing, such as moving objects in a scene~\cite{chen2021animatable}, modifying textures~\cite{xiang2021neutex}, and altering object shape~\cite{yang2022neumesh}. However, these editing procedures still require extensive user input. While recent work has enabled NeRF editing with text prompts~\cite{haque2023instruct}, it struggles to achieve precise and high-quality editing due to a restricted diversity of instructions.
Consequently, further research is needed to develop \textit{easy-to-use} and \textit{accurate} 3D editing methods, enabling improved re-creation of existing 3D assets.

% Para 3: 
% 1. the input and output.
% 2. Two core parts (a. mesh: locating, disentanglement, local; b. 流程, 着重强调定位的方法，optimize local feature, no need to change global feature)
In this paper, we present \sysname{}, a framework that allows users to intuitively and conveniently modify neural fields using text prompts. As illustrated in Fig.~\ref{fig:teaser}, for a given scene represented by a neural field, \eg a dog or a complex outdoor environment, text descriptions can be used to achieve various object-centric editing, including re-texturing, object replacement, and object insertion, while simultaneously preserving irrelevant regions to the text prompts. 
This is made possible through two key designs in our method: (1) a mesh-based neural field representation, and (2) a stepwise framework that leverages pretrained diffusion models for 3D editing. 
Compared to an implicit representation, an explicit mesh-based neural field enables the efficient conversion of 2D editing masks into 3D editing regions through back projection, facilitating precise local editing by only modifying the masked regions. Additionally, the mesh representation disentangles geometry and texture, preventing unnecessary geometry deformation when only appearance changes are expected. Leveraging the advantages of the mesh representation, we propose a stepwise \textit{finetune-localization-optimization} framework that efficiently and accurately edits 3D scenes using simple text prompts, achieved by score distillation sampling within the masked region.
% distilling a pretrained text-to-image diffusion model.
% Specifically, we first finetune the pretrained diffusion model on the input scene to bind it with a unique identifier. \sysname{} further locate the relevant regions based on the observation that the weights of attention layers in diffusion models are strongly correlated with text tokens. Then only the masked region will be optimized with the Score Distillation Sampling loss to achieve desired editing effects.

% Para 4: The results of our paper and why it is better
We extensively evaluate \sysname{} on various synthetic and real-world scenes, including animals, human faces and outdoor scenes. Unlike methods that operate on the entire image, our editing approach enables precise local deformations while naturally preserving irrelevant areas. For example, in Fig.~\ref{fig:teaser}, only the dog's mouth is modified when holding a rose in its mouth. Furthermore, as the edit can be accomplished with a simple text prompt, the procedure is user-friendly and significantly simplifies the editing of neural fields, showing its great potential for practical applications.
Both qualitative and quantitative comparisons demonstrate the superiority of \sysname{} over previous methods in terms of editing precision, visual fidelity and user satisfaction.
% Benefiting from the mesh representation and three-stage pipeline, our method can effectively perform targeted 3D editing based on text prompts, generating high-quality textures and shapes while maintaining robust 3D consistency. With only text prompts needed, the editing procedure is user-friendly and substantially alleviates the difficulty of editing neural fields, 

The contributions of this paper can be summarized as follows: 
(1) We introduce a novel framework for text-guided 3D scene editing, which achieves highly realistic editing results for a wide range of real-world scenes; (2) We propose to use a mesh-based neural field to enable local modification of the scene and decouple texture and geometric features for flexible editing; (3) We devise a stepwise editing framework that first identifies the specific regions requiring editing according to text prompts and then performs modifications exclusively within the selected regions. This systematic procedure ensures precise 3D editing while minimizing unnecessary modifications in irrelevant regions.

\section{Related Works}

\subsection{Text-guided image generation and editing}
% Reed et al.~\cite{reed2016generative} proposed to generate images from text input based on Generative Adversarial Network (GAN). Since then, GAN has become the dominant model for text-to-image generation and led to many remarkable works~\cite{zhang2017stackgan, xu2018attngan}. 
The denoising diffusion probabilistic model~\cite{ho2020denoising, song2020denoising} has drawn great attention for its ability to generate high-quality images. 
% To produce semantically consistent images given natural language text prompts from users, pre-trained large language models (LLMs)~\cite{radford2021learning, raffel2020exploring} were introduced into diffusion models to encode text prompts. 
Later, diffusion models~\cite{ramesh2022hierarchical, saharia2022photorealistic, rombach2022high} trained on large-scale image-text paired datasets demonstrated astonishing performance in understanding complex semantics from text prompts (including nouns, adjectives, verbs, etc.) and generating corresponding high-quality images. 
% However, diffusion models also suffer from slow sampling images speed. 
% To address this issue, Song et al.~\cite{song2020denoising} replaced the Markovian forward process used in diffusion models~\cite{ho2020denoising} with a non-Markovian one, which accelerated the sampling speed. 
% Moreover, to further improve the sampling efficiency and image resolution, Robin et al.~\cite{rombach2022high} proposed to diffuse in the low-resolution latent space and decode it into a high-resolution image. 
% Due to the success of diffusion models in image generation, many works also attempted to apply them to image editing.
Due to the rich semantics and high controllability of pretrained text-to-image diffusion models, a series of studies~\cite{couairon2022diffedit,kawar2022imagic,hertz2022prompt,avrahami2022blended} have employed them to text-guided image editing. 
% These methods achieve impressive progress in editing the attributes of objects, such as color, shape, and style, by controlling the diffusion process with text prompts or additional masks~\cite{huk2022shape}, exemplars~\cite{yang2022paint} et al. 
Most related to our work is subject-driven generation with text-to-image diffusion models~\cite{ruiz2022dreambooth, gal2022image}, which enables users to personalize their image generation for specific subjects and concepts given. 
DreamBooth~\cite{ruiz2022dreambooth} expands the language-vision dictionary using rare tokens and finetunes the model with a preservation loss for regularization. 
Similarly, Textual Inversion~\cite{gal2022image} optimizes a new ``word" in the embedding space of the pre-trained diffusion model to represent the input objects. 
These works address the task of editing specific images or generating images with novel concepts, but it is non-trivial to extend these 2D methods to 3D.

\subsection{Text-to-3D generation}

With the development of text-to-image generation models, there has been a growing interest in text-to-3D generation. 
Some works use the CLIP model to optimize mesh ~\cite{michel2022text2mesh, chen2022tango, mohammad2022clip} or neural fields ~\cite{jain2022zero}.
The seminar work DreamFusion~\cite{poole2022dreamfusion} first proposes  score distillation sampling (SDS) loss to distill the knowledge in pretrained 2D Text-to-Image diffusion models for text-to-3D generation. 
A series of works~\cite{metzer2022latent, lin2022magic3d, chen2023fantasia3d, raj2023dreambooth3d} based on SDS loss, further improve the generation results by introducing geometry prior or changing 3D representation.
% \revise{LatentNeRF~\cite{metzer2022latent} uses a Sketch-Shape to define base geometry and optimizes neural fields in latent space.}
Score Jacobian Chaining~\cite{wang2022score} arrives at a similar training objective from the perspective of approximating 3D score with the 2D score. 
% Magic3D~\cite{lin2022magic3d} improves Dreamfusion by proposing to use coarse representation as the initialization and optimize a textured 3D mesh model. Fantasia3D~\cite{chen2023fantasia3d} learns to disentangle geometry and appearance, supporting relighting and editing of the generated 3D assets.
% \revise{DreamBooth3D~\cite{raj2023dreambooth3d} provides a solution for generating personalized 3d assets by combining DreamFusion and Dreambooth.}
%Many other works investigate how to incorporate additional input images into the optimization pipeline, e.g. by applying a reconstruction loss and using predicted monocular depth~\cite{deng2022nerdi} or object mask~\cite{melas2023realfusion}.
However, all these methods lack the ability to edit existing 3D scenes. One of the main reasons is the difficulty in fully aligning an existing 3D scene with text, resulting in these methods tending to generate a new scene and breaking the consistency before and after editing.
To overcome this limitation, we propose a novel text-guided 3D editing framework that can edit an existing 3D scene based on text prompts.

\subsection{Neural Field Editing}
Editing neural fields is inherently difficult due to its entangled shape and appearance.
EditNeRF~\cite{liu2021editing} is the first work to support editing the shape and color of neural fields conditioned on latent codes. 
Some works~\cite{wang2022clip,wang2023nerf,gao2023textdeformer,bao2023sine} further leverage a CLIP model to allow editing with text prompts or reference images.
Another line of work uses pre-defined template models or skeletons to support re-posing or re-rendering~\cite{peng2021neural, noguchi2021neural}, but is constrained in a specific category. 3D editing can also be achieved by combining 2D image manipulation such as inpainting with neural fields training~\cite{liu2022nerf, kobayashi2022decomposing}. Geometry-based methods~\cite{yuan2022nerf, yang2022neumesh, xu2022deforming, li2022climatenerf} export neural fields to mesh and synchronize the deformation of the mesh back to implicit fields. 
% ClimateNeRF~\cite{li2022climatenerf} also extracts explicit geometry from a neural field and simulates weather changes to the scene with physical simulation. SINE~\cite{bao2023sine} is a image-based 3D editing approach, which edit a 3D scene with a single-view image.
TEXTure~\cite{richardson2023texture} uses a text prompt to generate the textures of the mesh using an iterative diffusion-based process.

The most similar work to ours is Instruct-NeRF2NeRF~\cite{haque2023instruct} and Vox-E~\cite{sella2023vox}, which edit a neural field freely text prompts. 
Instruct-NeRF2NeRF uses image-based diffusion model~\cite{brooks2022instructpix2pix} to edit the input image with instructions for optimizing the neural field. Nonetheless, since it manipulates the entire image, usually undesired regions will also be changed.
Vox-E adopts SDS loss and performs local editing in 3D space by 2D cross-attention maps. However, due to the constraints inherent of Vox-E's volumetric representation, the editing quality of real scenes remains suboptimal.

\section{Background}
{\textbf{Optimizing Neural Fields with SDS Loss.}}
DreamFusion~\cite{poole2022dreamfusion} proposed the score distillation sampling (SDS) loss to distill the priors Text-to-Image (T2I) diffusion models for 3D generation.
% for optimizing NeRF through text-based guidance with the aid of T2I diffusion models. 
It first adds random Gaussian noise at level $t$ to a random rendered view $\hat{I}$ to get $\hat{I}_{t}$. The pretrained diffusion model $\phi$ is used to predict the added noise given $\hat{I}_{t}$ and the input text condition $y$. 
% The diffusion model, equipped with a learnable denoising function $\epsilon_{\phi}(\hat{I}_{t};y,t)$, predicts the added noise. 
The SDS loss is calculated as the per-pixel gradient as follows:
\begin{equation}
  \nabla_{\theta}\mathcal{L}_{SDS}(\phi, \hat{I}=g(\theta))=\mathbb{E}_{\epsilon,t}\bigg[w(t)(\epsilon_{\phi}(\hat{I}_{t};y,t)-\epsilon)\frac{\partial \hat{I}}{\partial \theta } \bigg],
\end{equation}
where $w(t)$ is a weighting function that depends on noise level $t$, $\theta$ is the parameters of neural field and $g$ is the rendering process.
During training, the diffusion model is frozen and gradients are back-propagated to $\theta$, enforcing the neural field's renderings to resemble the images generated by the diffusion model with the text condition $y$.
% SDS loss pushes the noised rendered image towards a lower energy state of the T2I diffusion model, taking the text embedding $y$ as the condition.

\vspace{0.1cm}
\noindent{\textbf{DreamBooth}}~\cite{ruiz2022dreambooth} is a subject-driven image generation method based on T2I models. Given a few images of the same subject, DreamBooth embeds the subject into a T2I diffusion model by binding it to a unique identifier (denoted as $*$). It uses an L2 reconstruction loss to fine-tune the diffusion model on the input images and a class prior-preserving loss to prevent overfitting. The details of its training can be found in Ruiz et al~\shortcite{ruiz2022dreambooth}. In this paper, we also adopt DreamBooth to fine-tune the T2I diffusion models for expressing a specific scene.

% However, due to the complexity and diversity of real-world scenes, the image edited by DreamBooth tends to lose scene features such as the background, object shapes, or texture details. Similarly, using diffusion models fine-tuned by DreamBooth to guide the editing of a neural field also fails to maintain the consistency of the scene. 

\begin{figure*}
  \includegraphics[width=0.9\textwidth]{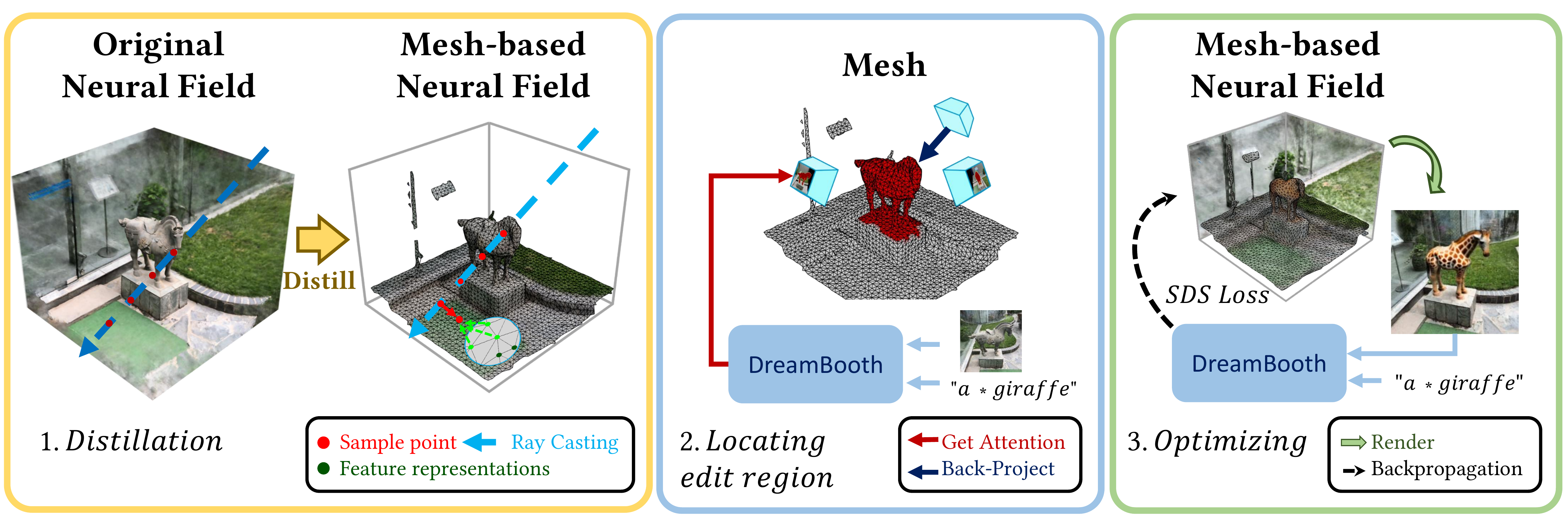}
  \caption{The overview of our method. Our method edits a 3D scene by optimizing an existing neural field to conform with a target text prompt.
  % takes an existing NeRF and a text prompt as input, ultimately producing an optimized NeRF that aligns with the text prompt. 
  The editing process involves three steps: (1) The original neural field is distilled into a mesh-based one. (2) Based on the text prompts, our method automatically identifies the editing region of the mesh-based neural field. (3) Our method utilizes the SDS loss to optimize the color feature $f_c$, geometry feature $f_g$, and vertex positions $v$ of the editing region, thereby altering the texture and geometry of the respective region. Best viewed in color.}
  \label{fig:overview}
\end{figure*}

\section{Method}

% view sample and 3D view back projection
% 场景assumption, 更多细节

% {\textbf{Problem setup.}}
\subsection{Overview}
The inputs of our method are a set of posed images of a 3D scene to be edited and a text prompt for editing. 
Our goal is to change the shape and appearance of the object of interest in the original 3D scene according to the text prompt. 
Fig.~\ref{fig_visual_compare} gives an example of turning a horse sculpture into a real giraffe. This task requires keeping the 3D contents irrelevant to the text prompt unchanged before and after editing. 
% To achieve this goal, we first learn a neural field using NeuS2~\cite{wang2022neus2} and then use a pretrained Stable Diffusion model ~\cite{rombach2022high} as the T2I diffusion model to guide the editing process. After the neural field of the scene is obtained, the T2I diffusion model is first fine-tuned by Dreambooth~\cite{ruiz2022dreambooth} to customize it to the 3D scene. 

The framework of \sysname{} is shown in Fig.~\ref{fig_visual_compare}, which consists of three stages. We first transform the original neural radiance field into a mesh-based neural field (Section~\ref{subsec:distill}), which enables us to achieve spatially-selective editing. In Section~\ref{subsec:locate}, we customize the T2I model to the input scene and use the cross-attention maps of it to locate the editing area in the 3D space according to the keywords in the text prompts. Finally, we edit the target object in the neural field under the control of text prompts through the T2I diffusion model (Section~\ref{subsec:optimize}).

\subsection{Distilling Neural Fields}
\label{subsec:distill}

% First of all, we distill mesh-based implicit fields (mesh-based NeRF) from the input NeRF.
Inspired by~\cite{yang2022neumesh}, we first learn a neural radiance field from input images and decompose it into many local implicit fields organized in an explicit mesh, where the mesh is extracted from the neural radiance field using marching cubes~\cite{lorensen1987marching}. 
Representing a scene as a mesh-based neural field introduces two benefits.
First, a mesh-based neural field enables precise editing of specific regions in the scene. 
The regions, such as background and irrelevant objects, can remain unchanged during editing by fixing the specific implicit fields. Second, the extracted mesh can explicitly represent the surface and outline of the objects in the scene. Compared with other explicit representations such as voxels~\cite{liu2020neural} and point clouds~\cite{ost2022neural}, it is more convenient to determine the range of editing area with mesh. Combining the attention scheme of the diffusion model, we further propose a method to automatically determine the editing area, which can accurately locate the editing area in the mesh according to the input text.

Specifically, after the neural radiance field is obtained, we adopt a teacher-student based  training framework to perform distillation, where the neural radiance field is taken as the teacher model to guide the student model, i.e., the mesh-based neural field. 
We define the mesh-based neural field by assigning each mesh vertex $\mathbf{v}$ a color feature ${f}_{c}$ and a geometry feature ${f}_{g}$ to represent the local shape and texture information near $\mathbf{v}$, respectively. During the volume rendering process, for a sampled point $x$, we first obtain the aggregated features ${\tilde{f}}_{c}$ and ${\tilde{f}}_{g}$ by interpolating the features of the top $K$ nearest vertices of $x$ weighted by the inverse distance ($\mathbf{v}_k -x$)~\cite{qi2017pointnet}:
\begin{equation}
  \tilde{{f}_{t}}(x)=\frac{ {\textstyle \sum_{k=1}^{K}}w_{k}{{f} }_{t,k}}{{\textstyle \sum_{k=1}^{K}}w_{k}} , w_{k}=\frac{1}{||\mathbf{v}_k -x|| } , t \in \{g, c\}
\end{equation}

Then, ${\tilde{f}}_{g}$ and ${\tilde{f}}_{c}$ are decoded to the s-density $s$ and color $c$ of $x$:
\begin{align}
s&=D_{G}({\tilde{f}}_{g}, {\tilde{h}}) , \quad c=D_{C}({\tilde{f}}_{c}, {\tilde{h}},\mathbf{d}, \nabla _{x}s)
\end{align}

where $D_{G}$ and $D_C$ are the geometry decoder and color decoder respectively, ${\tilde{h}}$ is the interpolated signed distance of $x$ to $\mathbf{v}_k$, $\mathbf{d}$ is the ray direction and $\nabla _{x}s$ is the gradient of s-density $s$ at point $x$. The framework of the network is shown in Fig.~\ref{fig:network}.

During the distillation process, we randomly sample rays $r$ in the scene and use the output of the teacher model given $r$ as the ground truth, including the rendered pixel color $\hat{C}(r)$, s-density $\hat{s}_i$ and point color $\hat{c}_i$ of each sampling point $x$ on this ray. 
The distillation loss is computed as:
\begin{equation}
  \mathcal{L}_{dis} = \sum_{r\in R}^{}\sum_{i\in N}^{} (\left \| \hat{s}_i -  s_i \right \| + \left \| \hat{c}_i -  c_i \right \|)+ 
 \sum_{r\in R}^{} \left \| \hat{C}(r)-C(r) \right \|_{2}^{2},
\end{equation}
where the volume rendering formulation of teacher and student models (\ie $\hat{C}$ and $C$) is the same as NeuS~\cite{wang2021neus}. Besides, we add Eikonal loss~\cite{gropp2020implicit} on the sampled points to regularize the norm of the spatial gradients with weight $\lambda_{reg}=0.01$ 
\begin{equation}
  \mathcal{L}_{reg} = \sum_{r\in R}^{}\sum_{i\in N}^{} \left \| \left \| \nabla _{x_i}s_i \right \|  -1 \right \|^{2}_{2}.
\end{equation}

% The final objective of distillation is formulated as:
% \begin{equation}
% \min_{f_c,f_g,D_C,D_G}  \mathcal{L}_{dis}  + \lambda_{reg}\mathcal{L}_{reg} .
% \end{equation}
% where $\lambda_{reg}$ is a weight between the distillation loss and Eikonal loss. 

In our framework, all camera pose sampling is based on the spherical coordinate system. 
We transform the target object to the origin and make the y-axis point upwards.
% which requires the transformation of the coordinate system of the input scene. 
% Leveraging mesh visualization tools such as MeshLab, we are able to obtain the transformation matrix by moving the coordinate origin to align with the target object and ensuring the y-axis is pointed upwards. 
We confine the scope of sampled views by setting the range of the elevation and azimuth angles in the following locating and optimizing step, thereby improving editing efficiency.

\subsection{Locating Editing Regions}
\label{subsec:locate}

As illustrated in the middle part of Fig~\ref{fig:overview}, given text prompts, \sysname{} first determines the target editing area in a rendered view. 
As a preparation step, we first fine-tune the Stable Diffusion model with DreamBooth with the sampled views, which adapts the model's knowledge to the specific scene.
Then, we utilize the fine-tuned diffusion model to obtain a 2D mask for each rendered view. Finally, we obtain the 3D editing region by back-projecting the masked target region from different views onto the mesh. 
% By fixing the implicit fields in the non-editing regions, we ensure that the background and other objects irrelevant to the text prompt remain unchanged during editing. 

The locating is motivated by the fact that cross-attention layers in T2I diffusion models control the relationship between the layout of the generated images and each word~\cite{hertz2022prompt}: $M=\mathrm{Softmax}({QK^{T}}/{\sqrt{q} } ),$
where $Q$ is the query features projected from the spatial features of the noisy image with dimension $q$, $K$ is the key matrix projected from the textual embedding, $M$ is the attention map that defines the weight of a token for each pixel. Therefore, $M$ indicates the probability that a pixel corresponds to a word in the text prompt and can be utilized to locate the editing area.
% . Therefore, we exploit the attention map in the cross-attention layers to locate the editing area in an image
%. The text prompts are first encoded into textual embedding $y$ using pre-trained text encoders (such as CLIP~\cite{radford2021learning}). Then, these embeddings are injected into cross-attention layers to control the output of the diffusion model:
% \begin{equation}
% M=\mathrm{Softmax}(\frac{QK^{T}}{\sqrt{q} } ) ,
% \end{equation}

\begin{figure*}
  \includegraphics[width=0.95\textwidth]{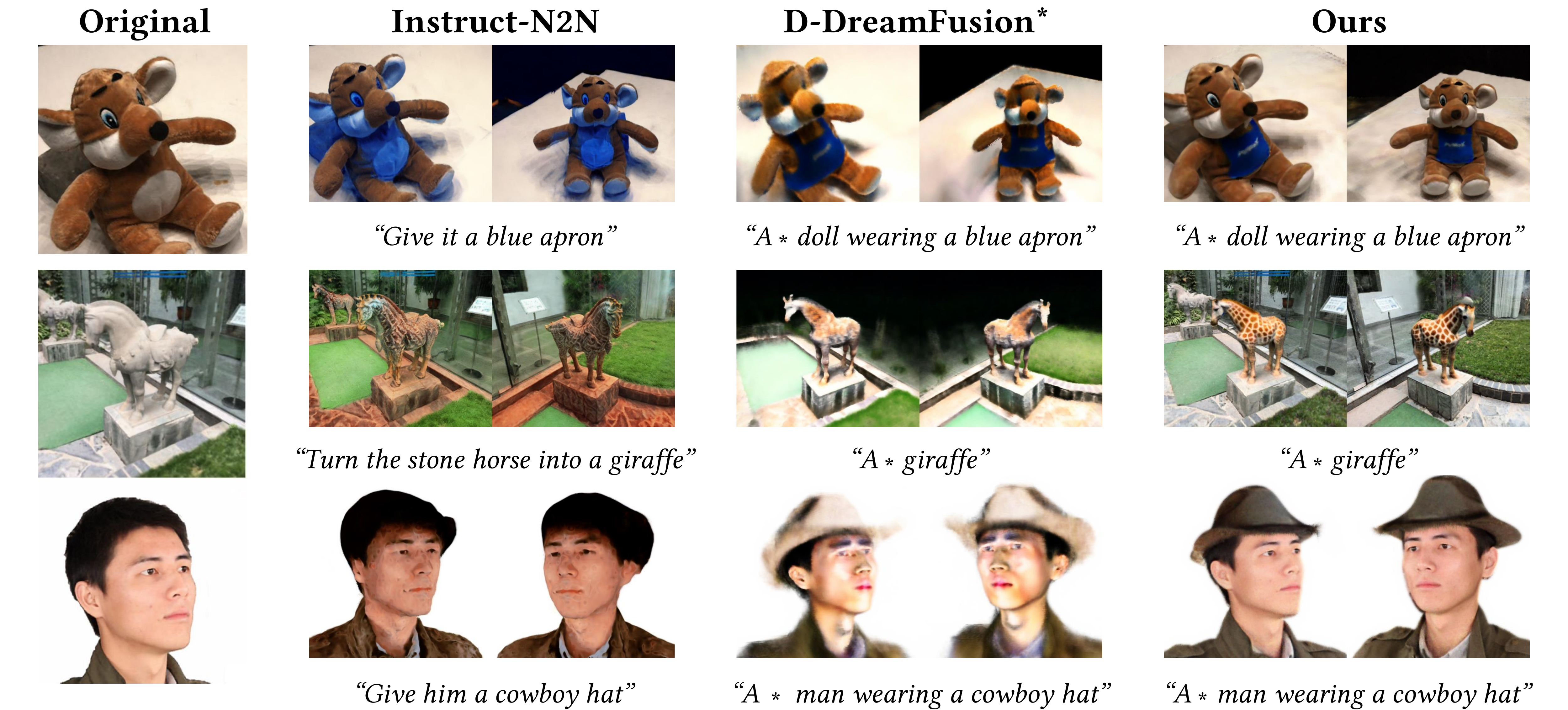}
  \caption{\textbf{Visual results} of our method compared with two baselines on three different scenes. The results clearly show that \sysname{} can precisely locate the relevant region, perform faithful editing to the text, and prevent undesirable modifications, which are difficult to be achieved by the baseline methods.}
  \label{fig_visual_compare}
\end{figure*}

Specifically, the noisy image $\hat{I}_{t}$ of a rendered view and the text prompt are fed into the diffusion model for denoising. We select the keyword that represents the intended editing results (e.g., "apron", "giraffe", "hat" as in Fig.~\ref{fig_visual_compare}) and extracts all its attention maps produced during the generation process. In practice, the backbone of the diffusion model usually consists of $L$ convolutional blocks, which are equipped with $H$ multi-headed attention layers~\cite{vaswani2017attention}. Therefore, after $T$ rounds of denoising, the final set of attention maps $\mathbf{M}$ can be represented as $\left \{ M_{t,l,h}\right \} $, where $t$, $l$, $h$ represent the index of the time step, convolution block, attention head, respectively. We resize all attention maps to the same resolution by bilinear interpolation and aggregate them to obtain the aggregated attention map $\overline{M}$. $\overline{M}$ are further normalized to [0,1] and binarized with a threshold $\tau=0.75$, where the area with a value of 1 is the editing area. We back-project all the pixels belonging to the editing area in the mask onto the mesh and mark the intersected mesh faces as the editing region. It is worth highlighting that the keywords are not restricted to the objects in the initial scene, as the attention maps of a keyword delineate regions in the generated image where the likelihood of keyword presence is highly probable. As shown in Fig.~\ref{fig_locating}), even though "sunglasses" is not part of the original scene, it remains feasible to identify the reasonable region on the scene mesh.

In this stage, we traverse all elevation and azimuth angles at 45° intervals within the scope of sampled views to ensure the coverage of all potential editing regions. 
Subsequently, we get the masks of all sampled views and back-project them onto the mesh. After merging the results of back-projection, we employ two steps to refine the masked region: (1) Discard: we discard the small pieces within the editing region where the number of faces is less than 10\% of the total projected area, which typically emerges from inaccuracy 2D masks (\eg masks larger than target object is projected outside the object); (2) Fill: we use breadth-first search to fill in the ``holes'' in the editing region, \ie a non-editing region surrounded by editing regions. Such "holes" usually come from occluded (\eg the bottom of a horse) or concave areas. By integrating these regions into the editing area, we enhance the completeness of the editing area.
We denote the final editing region as $\mathbf{V}=\left \{  v_{e}\right \}^{E}_{e=1}$.

\subsection{Optimizing Editing Regions}
\label{subsec:optimize}

In this step, we adopt the SDS Loss from DreamFusion~\cite{poole2022dreamfusion} to guide the optimization of the editing region in the neural field with the T2I diffusion model, making the scene conforms to the text prompt. By feeding random rendered views and the text prompt to the T2I diffusion model, we calculate the SDS Loss and backpropagate the gradients to the neural field.
% , allowing the rendering in the editing region to look like the image generated by the T2I model for the given text prompt. 
Since the Imagen~\cite{saharia2022photorealistic} in DreamFusion is proprietary, we compute the SDS Loss in the latent space with Stable Diffusion~\cite{rombach2022high} as follows:
\begin{equation}
\nabla_{\omega }\mathcal{L}_{SDS}(\phi, g(\omega)) = \mathbb{E}_{\epsilon,t}\bigg[w(t)(\epsilon_{\phi}(z_{t};y,t)-\epsilon)\frac{\partial z}{\overset{}{\partial} \hat{I} } \frac{\partial \hat{I}}{\overset{}{\partial} \omega   } \bigg],
\label{equ:oursds}
\end{equation}
where $\omega = \{f_{g,k}, f_{c,k}, \mathbf{v}_k\}_{k}$ are the set of geometry features, color features and positions for all mesh vertices in $\mathbf{V}$, $z_{t}$ denotes the noisy latent, and $z$ is the original latent generated by the encoder of the Stable Diffusion model.

We can see from Equation \ref{equ:oursds} that during training, apart from optimization of the color feature $f_{c}$ and geometry feature $f_{g}$ of the vertices in the editing region, the positions of the vertices are also included. This implies that the structure of the mesh is also dynamically adjusted during the optimization, which is a critical part of our approach. 
In local implicit fields, geometry features mainly represent shape details near the vertices. 
The smoothness of the object's surface will be disrupted if there are significant changes in the s-density of the points situated away from the vertices.
Hence, we propose a complementary optimization approach, which simultaneously optimizes the vertex position and geometry features. The optimization of the vertex position ensures that the overall shape of the mesh conforms to the text prompt, while the optimization of the geometry features refines the local geometry of the object. This optimization approach enables \sysname{} to generate complex shapes, such as rose petals. Our ablation study in Section \ref{section:AS} demonstrates the necessity of the joint optimization of the vertex position and geometry features.

To maintain a smooth surface and encourage natural deformation during vertex position optimization, we introduce widely-used mesh regularization terms, including the Laplacian loss and ARAP (as-rigid-as-possible) loss~\cite{sumner2007embedded}:
\begin{align}
\mathcal{L}_{lap} &= \frac{1}{E} \sum_{i=1}^{E} \begin{Vmatrix}
\mathbf{v}_{i}-  \frac{1}{|N_{i}|} \sum_{j\in N_{i}}^{}  \mathbf{v}_{j} 
\end{Vmatrix} ^{2}, \\
\mathcal{L}_{ARAP} &= \sum_{i=1}^{E}\sum_{j\in N_{i}}^{} \left | ||\mathbf{v}_{i}-\mathbf{v}_{j}||_{2}- ||\mathbf{v}'_{i}- \mathbf{v}'_{j}||_{2} \right | ,
\end{align}
where $N_{i}$ is the set of one-ring neighbours for vertex ${v}_{i}$, ${v}'$ indicates the vertex position in the last iteration.
We set $\lambda_{lap}=10^{-4}$ and $\lambda_{ARAP}=10^{-4}$ to balance them respectively.

We perform both the SDS Loss and mesh regularization terms during optimization in each iteration. We found that optimizing the SDS and regularization terms separately achieves better results empirically. Given a rendered view, we first optimize $f_{c}$, $f_{g}$, $\mathbf{v}$ of the editing region with the SDS loss. Then, $f_{c}$ and $f_{g}$ are fixed, and only $\mathbf{v}$ is optimized with the mesh regularization terms.

% and the objective of mesh regularization is $\min_{\mathbf{v}} (\lambda_{lap}\mathcal{L}_{lap}  + \lambda_{ARAP}\mathcal{L}_{ARAP})$.

\section{Experiments}
% In this section, we demonstrate experimental results on a variety of real-world scenes, including both qualitative and quantitative comparisons against baseline methods. 
% In this section, we demonstrate experimental results on a variety of real-world scenes, including single-object scenes with pure background, human faces, and indoor as well as outdoor scenes with complex backgrounds.
% To evaluate the efficacy of our approach, we perform both qualitative and quantitative comparisons against baseline methods. 
% First of all, we provide a detailed description of the experiment setting.

\subsection{Experimental Setup}

\vspace{0.1cm}
\noindent{\textbf{Dataset}}.
To verify the effectiveness of our method in various scenes, we select six scenes with different levels of complexity from four datasets: DTU~\cite{jensen2014large}, BlendedMVS~\cite{yao2020blendedmvs}, Co3D~\cite{reizenstein2021common}, and GL3D~\cite{shen2019matchable}. These scenes include objects in simple backgrounds, human faces, and outdoor scenes with complex backgrounds. We use high-resolution images and the corresponding camera poses from the respective datasets to learn the original neural fields. Then, we edit the original scenes based on text prompts.

\vspace{0.1cm}
\noindent{\textbf{Baselines}}.
We compare with three baselines. 
%(1) \dreamfusion{}: the original paper uses a proprietary Imagen diffusion model that is publicly inaccessible, thus, we adopt an unofficial implementation based on the Stable Diffusion model, named as \dreamfusion{}~\cite{stable-dreamfusion}.
(1) \dreambooth3d{}: as pointed out by \nerf2nerf{},  DreamFusion fails to edit a neural field due to the difficulty of finding an exact textual description that matches a scene. To learn a better neural representation of a specific scene, we combine Stable-DreamFusion with DreamBooth~\cite{ruiz2022dreambooth} as another baseline. 
(2) Instruct-NeRF2NeRF (\nerf2nerf{}): we also compare with a recent work Instruct-NeRF2NeRF and use the text instructions provided by the paper~\cite{haque2023instruct} to edit a 3D scene.
(3) NeRF-Art~\cite{wang2023nerf}: Since NeRF-Art only supports stylized editing, we compare it in the stylization task.

\vspace{0.1cm}
\noindent{\textbf{Evaluation Criteria}}.
Following~\cite{haque2023instruct}, we use the CLIP Text-Image directional similarity to evaluate the degree of alignment between the change in both the images and text prompts and its detailed definition can be found in~\cite{gal2022stylegan}. For each editing result, we uniformly sample 50 viewpoints around the editing region and take the mean value as the result.
Since the CLIP directional similarity can only approximately evaluate the editing quality, we additionally conduct user studies to obtain human evaluations. We distribute 50 copies of questionnaires, presenting rotation video results of all methods side by side and asking users to choose the best editing result. The voting rates are calculated for each method.
We compare our method with the aforementioned baselines in four selected scenes, covering a total of 20 distinct editing operations. We exclude NeRF-Art in the quantitative comparison due to it only supports stylized editing.

\vspace{0.1cm}
\noindent{\textbf{Implementation Details}}. In our experiments, we adopt NeuS to learn the original neural field. The training parameters can be found in~\cite{wang2021neus}.
As for the diffusion model, we use the public pretrained Stable Diffusion model V2.
For each original neural field, we use the rendered images from the locating step, applying DreamBooth to fine-tune the Stable Diffusion model over 500 iterations.
% The Stable Diffusion model and DreamBooth are both implemented in HuggingFace Diffusers.
In the distilling step, we use the Adam optimizer with $lr=10^{-4}$ to optimize the local fields for 100K iterations.
In the optimizing step, the size of the rendered images is gradually increased from 96$\times$96 to 192$\times$192. We set the Adam optimizer with $lr=10^{-2}$ to optimize the $f_{c}, f_{g}, \mathbf{v}$ of vertices in the editing region for 2K iterations.
We implement our editing framework in Pytorch.

\subsection{Qualitative Results}

\textbf{Results of Editing 3D Scenes.}
We provide qualitative results of our method in Fig.\ref{fig:teaser} and Fig.~\ref{fig:gallery1}. Results demonstrate that our method can effectively perform targeted editing of neural fields in various scenes. As depicted in the middle row of Fig.\ref{fig:teaser}, even in complex scenes such as outdoor gardens, our method can accurately determine the horse sculpture as the editing region, subsequently turning it into a deer or giraffe with high-quality textures and geometry. Moreover, our method is capable of local editing, such as wearing sunglasses for the dog in the bottom of Fig.~\ref{fig:teaser}.
Notably, as shown in Fig.~\ref{fig:fig_mesh}, the editing results produced by our method demonstrate excellent consistency in 3D geometry, as can be intuitively observed in the extracted mesh.

Fig.\ref{fig_visual_compare} presents a comparison of the results of our method with baselines. 
\nerf2nerf{} has difficulties in executing abstract operations (e.g. give an apron to a doll) and generates suboptimal results in some scenes. This is largely attributed to the fact that the Instruct-Pix2Pix model is not always reliable, and it operates on the full image. 
Therefore, \nerf2nerf{} changes the entire scene and may underperform when executing the instructions beyond the Instruct-Pix2Pix training set.
The DreamBooth finetuning in \dreambooth3d{} enables the T2I diffusion model to roughly learn the representation of the input objects, such as the toy in the first row and the man in the third. However, due to the complexity and diversity of real-world scenes, \dreambooth3d{} cannot accurately represent a specific scene, leading the irrelevant regions of the scenes edited by \dreambooth3d{} to change significantly, such as the deformation of the doll in the first row, the background in the second row. 
Moreover, all compared baselines can not guarantee the consistency of the scenes before and after editing in complex scenes (such as the garden in the second row), and their editing process may change the entire scene. 
In contrast, our method has more details and faithfully generates the content of the text prompts, while successfully maintaining the consistency of the input objects and scenes before and after editing.

\vspace{0.05cm}
\noindent{\textbf{Results of stylization task.}}
As shown in Fig.\ref{fig_stylization}, we compare our method with NeRF-Art and \nerf2nerf{}. In this task, we omit the locating step to stylize the whole scene. Since stylization editing is a subjective task, we only provide the qualitative results as a reference and do not conduct quantitative analysis.

\vspace{0.05cm}
\noindent{\textbf{Results of locating editing region.}}
In Fig.\ref{fig_locating}, we also show our method's results of locating editing regions. We can see that our method can locate reasonable editing regions precisely.

\begin{table}[t]
  \caption{Results of the CLIP Text-Image Direction Loss and user studies.
  }
  \label{tab:Quantitative}
  \begin{tabular}{lcc}
    \toprule
    Method  & \makecell{CLIP Text-Image \\ Direction Similarity $\uparrow $}  & \makecell{Editing performance \\ voting percentage $\uparrow $ } \\
    \midrule
    % \dreamfusion{} & 10.72 & 1.6\% \\
    \dreambooth3d{} & 12.43  & 12.1\% \\
    \nerf2nerf{} & 10.86  & 6.8\% \\
    Ours & \textbf{18.49} & \textbf{\ourvote{}\%} \\
    \bottomrule
  \end{tabular}
\end{table}

\subsection{Quantitative Results}

In Table \ref{tab:Quantitative}, we present the results of the CLIP text-to-image directional loss. The results clearly demonstrate that our method achieves significantly higher scores, indicating that our method generates shapes and textures that are clearer and more aligned with the edited text prompts. 
Additionally, our method receives over \ourvote{}\% of the votes, surpassing the other methods by a significant margin. This further demonstrates \sysname{} can achieve much higher user satisfaction across various scenes.

\subsection{Ablation Study}
\label{section:AS}

\begin{figure}
  \includegraphics[width=0.9\columnwidth]{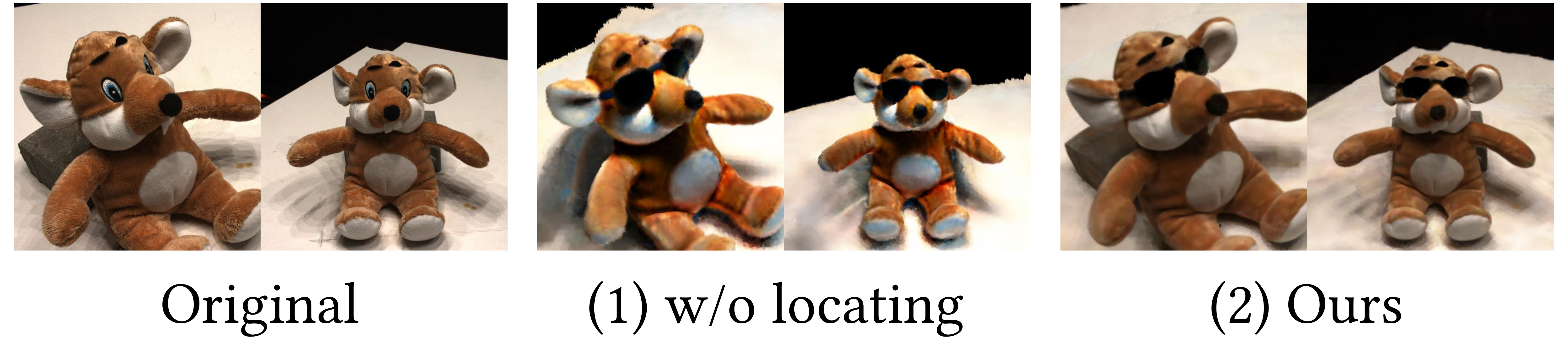}
  \caption{Ablation study of locating step. Editing without the locating step will deform the doll, breaking the consistency of the object.}
  \label{fig_ablation_locating}
\end{figure}

\textbf{Effectiveness of locating step.}
To demonstrate the necessity of locating step, we design two variants: (1) w$/$o locating: We omit the locating step and optimize all local implicit fields on the mesh. (3) Our method: we determine the editing region through locating step, and fix the non-editing region in optimization. 
As illustrated in Fig.\ref{fig_ablation_locating} (1), editing without the locating step will inadvertently change irrelevant regions of the scene, such as shortening the doll's arm, which destroys the consistency of the object. In contrast, the locating step allows our framework to optimize exclusively the region of interest. 
% This stark contrast fully demonstrates the importance of the locating step.

% \vspace{0.05cm}
\noindent{\textbf{Effectiveness of optimizing approach.}}
To evaluate whether our optimizing approach can generate more detailed 3D shapes during optimization, we ablate with three variants of \sysname{} as follows: (1) Fixing $\mathbf{v}$: fixing the mesh structure during the updating process, only optimizing the geometry features. (2) Fixing$f_{g}$: only changing the mesh structure without optimizing the geometry feature. (3) Our method: $\mathbf{v}$ and $f_{g}$ are optimized simultaneously. We select a challenging scene to evaluate: \textit{generating a rose on a cup}. 

\begin{figure}
  \includegraphics[width=0.9\columnwidth]{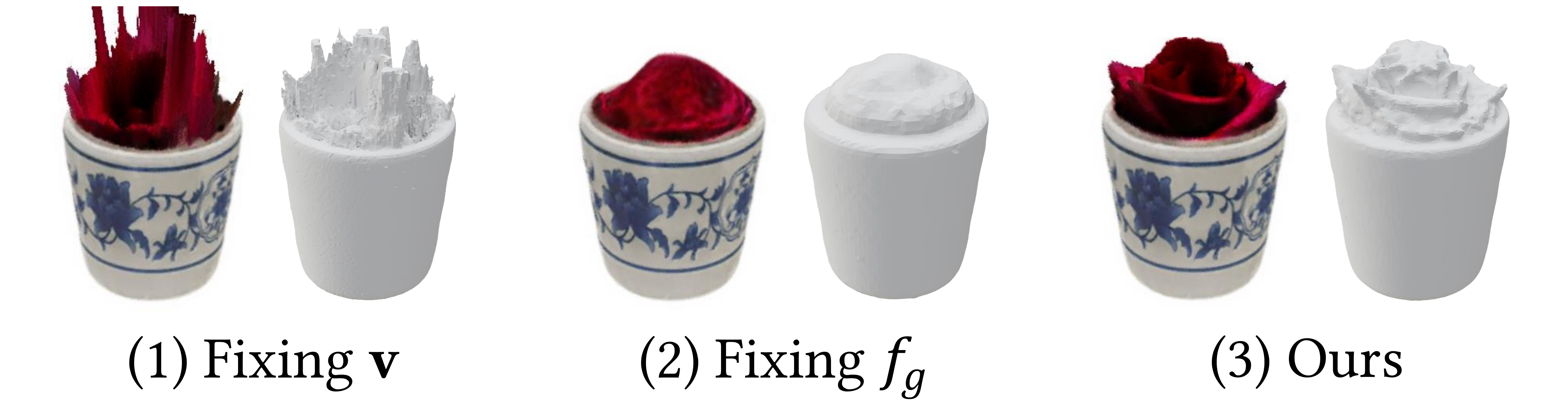}
  \caption{Ablation study of optimizing approach. Obviously, simultaneously optimizes both geometry features and vertex positions (Ours) and generates red roses with more detailed and realistic 3D shapes.}
  \label{fig_ablation}
\end{figure}

We present the rendered images of the generated results and the extracted 3D shape using the marching cubes algorithm in Fig.~\ref{fig_ablation}.  Fig.\ref{fig_ablation} (1) displays the rose generated by fixing vertex positions, which are full of spikes. This is because, in regions far from the mesh surface, constraining the smoothness of the s-density of the sampling points across implicit fields is quite challenging. Fixing geometry features, as shown in Fig.\ref{fig_ablation} (2), can generate a rough shape but lacks details. In contrast, our method simultaneously optimizes both the geometric features and vertex positions, which eliminates the spikes as well as generates more detailed buds and petals. 
% This demonstrates the ability of our method to generate detailed 3D shapes.

% \begin{figure}
%   \includegraphics[width=\columnwidth]{pic/limitation.pdf}
%   \caption{As other SDS-based 3D generation methods, our method may also suffer the Janus problem, where the edited doll has Iron Man's face on both the front and back.}
%   \label{fig_limitations}
% \end{figure}
% \subsection{Limitations}

% In BlendedMVS and GL3D datase with complex background, following~\cite{zhang2020nerf++}, we separately model foreground and background by the given foreground mask, and perform editing only on the foreground NeRF.

\section{Conclusion and Limitations}

In this paper, we present \sysname{}, a text-driven framework for editing 3D scenes represented by neural fields. Given a neural field and text prompts describing the desired edits, \sysname{} automatically identifies the editing region within the scene and modifies its geometry and texture accordingly. Experiments across a diverse range of scenes, including faces, objects, and large outdoor scenes, showcase the robust editing capabilities of \sysname{} to generate high-quality textures and shapes compared with other baselines while ensuring the edited scene remains consistent with the input text prompts.

Limitations of \sysname{} include the Janus problem, an issue inherited from DreamFusion, where the generated object appears as a front view from different viewpoints. Furthermore, \sysname{} does not directly model environmental lighting, which limits control over the lighting condition. While \sysname{} generally works well, due to the dependence of rendered views in editing, its performance may suffer in the presence of significant self-occlusions in the scene, consequently impacting the final synthesis results. Considering that NeuS faces difficulties in effectively reconstructing backgrounds in unbounded scenes, our current focus lies on object-centric editing in the foreground of the scene. In the future work, by combining recent unbounded real-world scene mesh reconstruction methods, such as BakedSDF~\cite{yariv2023bakedsdf}, our method can be extended to the whole scene editing.

% Future work could focus on adding an interface to enable users to control the view scope and presenting both the location area and the editing results effect interactively.
% In the future, we plan to introduce a lighting module that leverages rasterization based on the mesh structure to enhance shading effects. Moreover, we intend to construct an interactive visual editing interface, offering users the convenience of confirmation of the view scope and presenting both the location area and the editing results effect in real time.

\begin{acks}
This work was supported in part by the National Natural Science Foundation of China (NO.~62322608, ~61976250), in part by the Open Project Program of State Key Laboratory of Virtual Reality Technology and Systems, Beihang University (No.VRLAB2023A01), and in part by the Guangdong Basic and Applied Basic Research Foundation~(NO.~2020B1515020048).
\end{acks}

\clearpage

\appendix

\begin{figure*}[htbp]
	\centering
	\begin{minipage}{\columnwidth}
		 \includegraphics[width=\columnwidth]{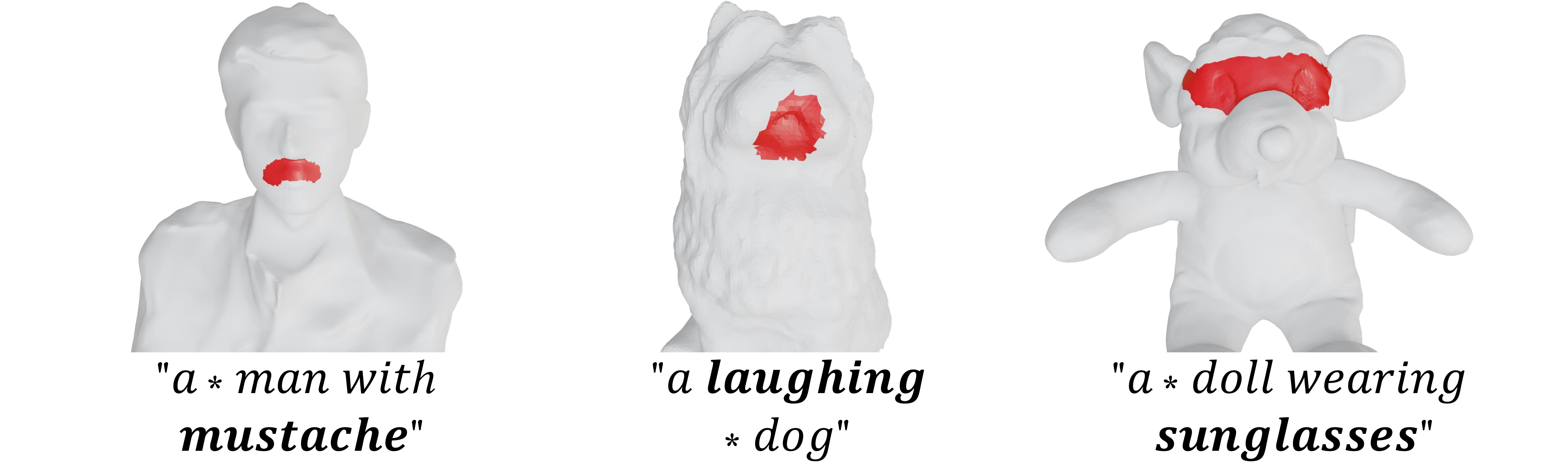}
  \caption{Visualization of the editing region, where the bold words indicate keywords and the red area on the mesh represents the editing region.}
  \label{fig_locating}
	\end{minipage}
	\qquad
	\begin{minipage}{\columnwidth}
		\centering
		 \includegraphics[width=\columnwidth]{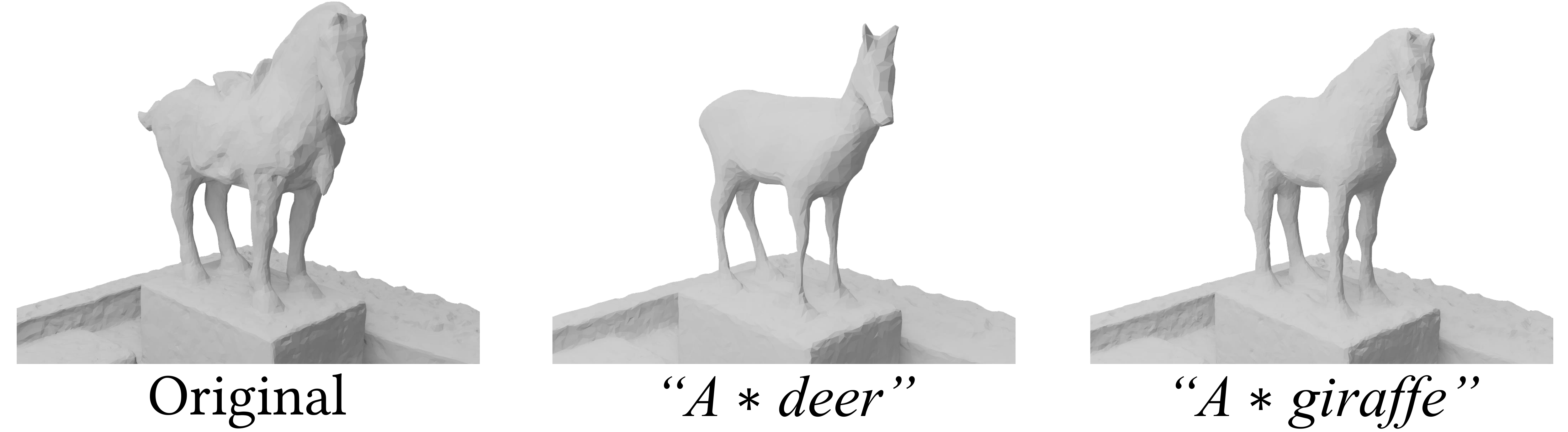}
  \caption{Visualization of the extracted mesh from our editing results.}
  \label{fig:fig_mesh}
	\end{minipage}
\end{figure*}

\begin{figure*}
  \includegraphics[width=0.62\textwidth]{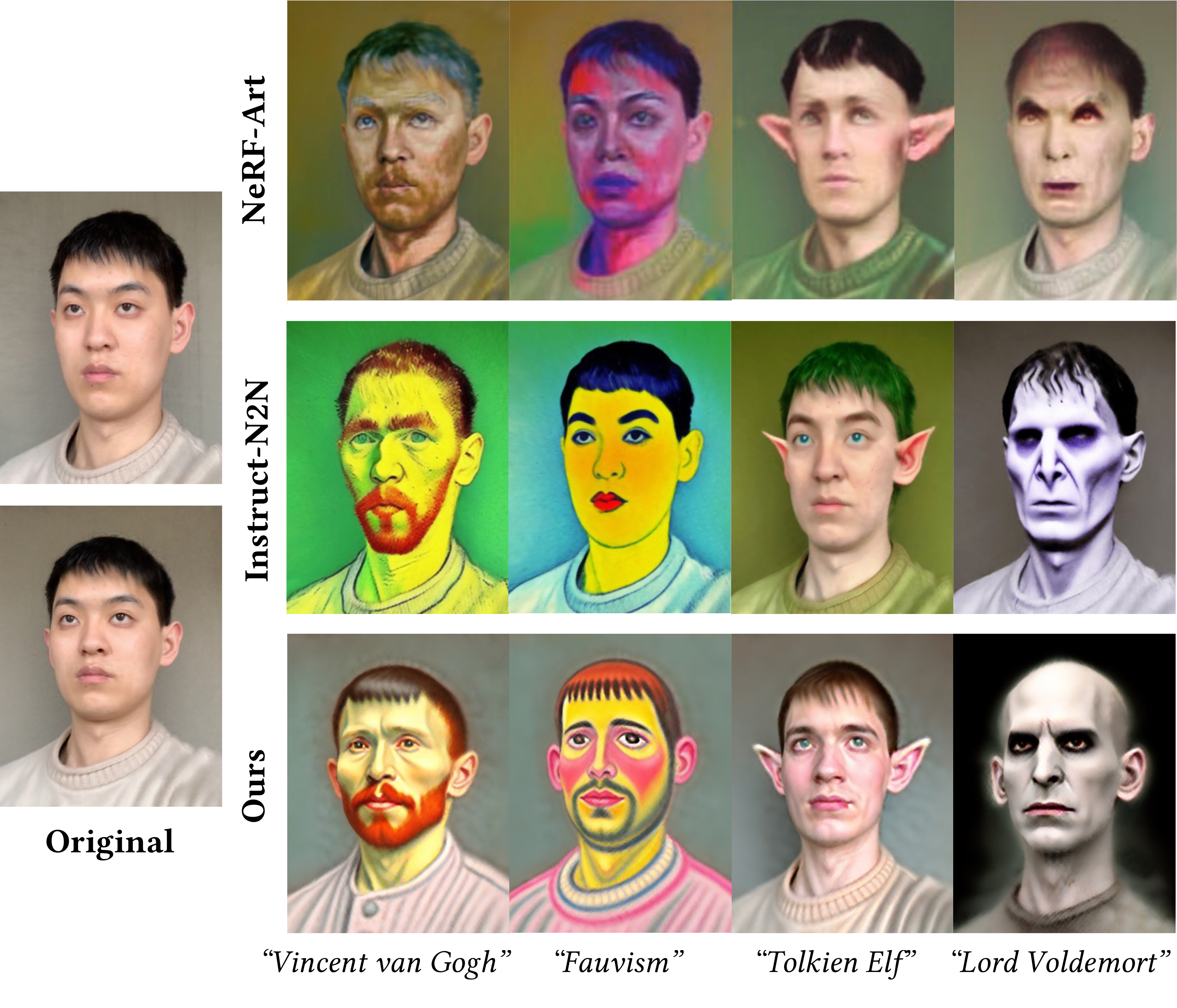}
 \caption{Visualization of the stylization editing results, we compare with NeRF-Art and Instruct-NeRF2NeRF.}
  \label{fig_stylization}
\end{figure*}

% \begin{figure}
%   \includegraphics[width=\columnwidth]{pic/locating.pdf}
%   \caption{Visualization of the editing region, where the bold words indicate keywords and the red area on the mesh represents the editing region.}
%   \label{fig_locating}
% \end{figure}

% \begin{figure}
%   \includegraphics[width=\columnwidth]{pic/vis_mesh.pdf}
%   \caption{Visualization of the extracted mesh from our editing results.}
%   \label{fig:fig_mesh}
% \end{figure}

\begin{figure*}
  \includegraphics[width=0.7\textwidth]{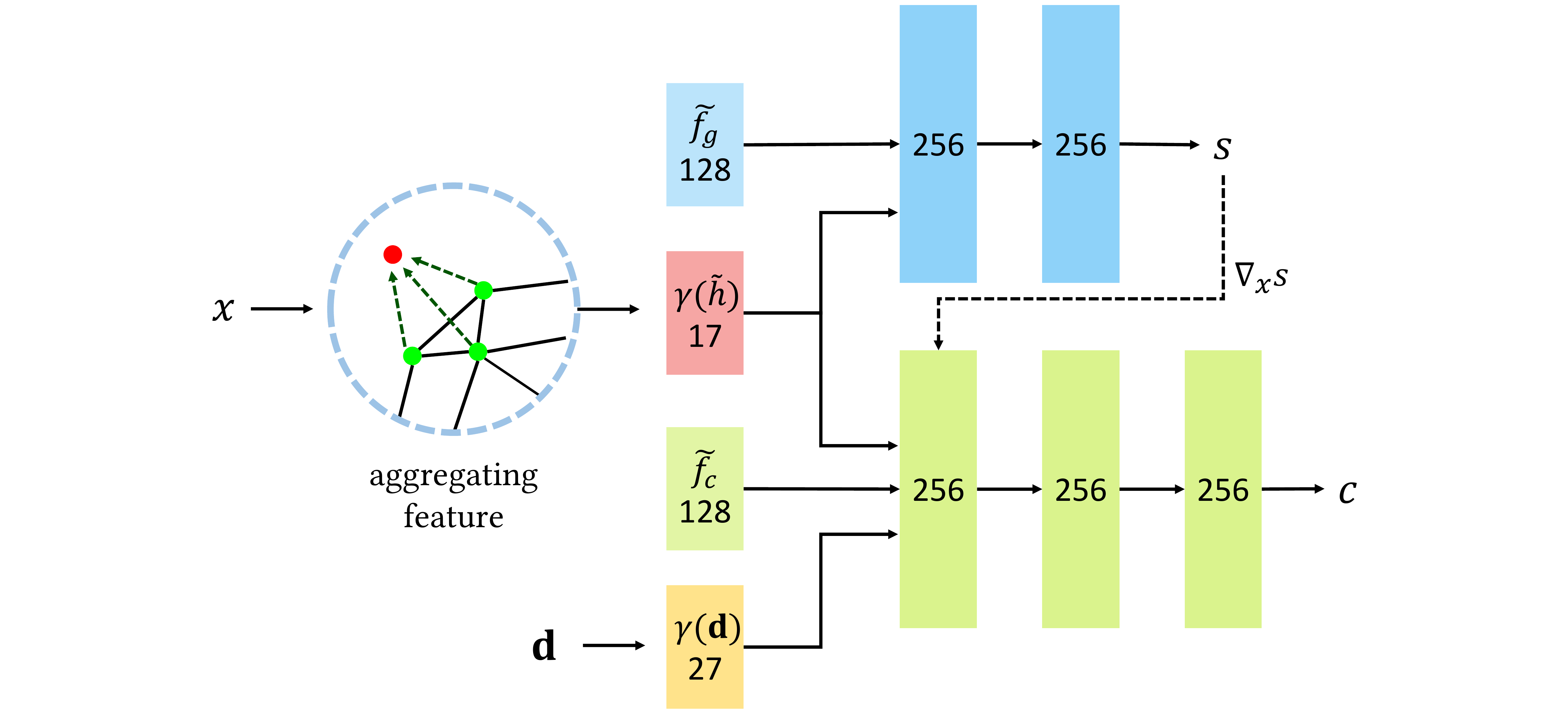}
  \caption{The network of the mesh-based neural fields. It takes the sampled point $x$ and the ray direction $\mathbf{d}$ as input, output the s-density $s$ and color $c$.  $\gamma (\cdot )$ denotes positional encoding adopted in NeRF~\cite{mildenhall2021nerf}.}
  \label{fig:network}
\end{figure*}

\begin{figure*}
  \includegraphics[width=\textwidth]{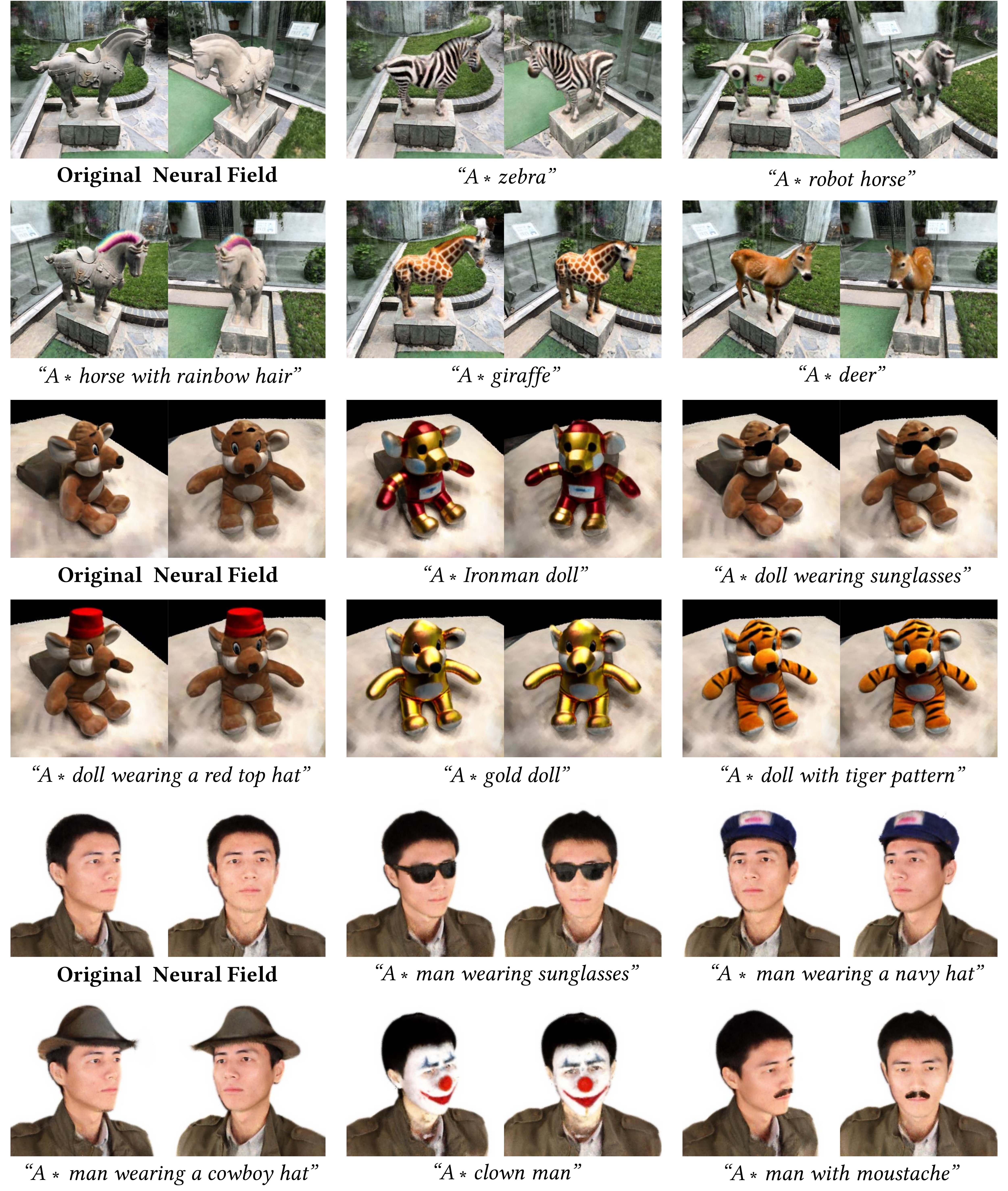}
  \caption{More editing results.}
  \label{fig:gallery1}
\end{figure*}

% \begin{figure*}
%   \includegraphics[width=\textwidth]{pic/g2.pdf}
%   \caption{More editing results.}
%   \label{fig:gallery2}
% \end{figure*}

\clearpage

\bibliographystyle{ACM-Reference-Format}
\bibliography{Dreameditor}

\end{document}